\documentclass[english,times,singlecolumn,authoryear,oneside, a4paper,12pt]{ article}

\usepackage{framed,multirow}
\usepackage{amssymb}
\usepackage{latexsym}
\usepackage{natbib}
\usepackage{url}
\usepackage{xcolor}
\definecolor{newcolor}{rgb}{.8,.349,.1}

\usepackage[pagebackref=true,breaklinks=true,letterpaper=true,colorlinks,bookmarks=false]{hyperref}
\usepackage{times}
\usepackage{epsfig}
\usepackage{graphicx}
\usepackage{amsmath}
\usepackage{algorithm}
\usepackage[noend]{algpseudocode}
\usepackage{authblk}
\usepackage{booktabs,caption,fixltx2e}
\usepackage[flushleft]{threeparttable}
\usepackage[T1]{fontenc}
\usepackage{subfig}
\usepackage{array}
\usepackage{multirow}
\usepackage{booktabs}
\usepackage{babel}
\usepackage{xspace}

\usepackage{calc}
\newlength\myheight
\newlength\mydepth
\settototalheight\myheight{Xygp}
\makeatletter

\usepackage{url}
\usepackage{xcolor}
\definecolor{newcolor}{rgb}{.8,.349,.1}
\DeclareOldFontCommand{\bf}{\normalfont\bfseries}{\mathbf}
\DeclareOldFontCommand{\it}{\normalfont\itshape}{\mathit}
\DeclareOldFontCommand{\rm}{\normalfont\rmfamily}{\mathrm}

\begin{document}

\clearpage

\title{Deep Spectral Correspondence for Matching Disparate Image Pairs}

\small
\author[1]{\small Arun  CS Kumar\textcolor{green}{*}}
\author[2]{\:\:\:\: Shefali  Srivastava\textcolor{green}{*}} 
\author[3]{\:\:\:\:Anirban   Mukhopadhyay} 
\author[1]{\:\:\:\:Suchendra  M. Bhandarkar} 

\affil[1]{\footnotesize Department of Computer Science, The University of Georgia, Athens, GA 30602-7404, USA}
\affil[2]{Department of Information Technology, Netaji Subhas Institute of Technology, New Delhi, India}
\affil[3]{Department of Computer Science, Technische Universitat Darmstadt, Darmstadt, Germany \newline \footnotesize(\textcolor{green}{*} - Joint First Authors) }
\date{\footnotesize{\it (Under submission to Computer Vision and Image Understanding (CVIU) journal)}}

\maketitle
\section{Abstract}

% what
A novel, non-learning-based, saliency-aware, shape-cognizant correspondence determination technique is proposed for matching image pairs that are significantly disparate in nature. 
% problem description
Images in the real world often exhibit high degrees of variation in scale, orientation, viewpoint, illumination and affine projection parameters, and are often accompanied by the presence of textureless regions and complete or partial occlusion of scene objects. The above conditions confound most correspondence determination techniques by rendering impractical the use of global contour-based descriptors or local pixel-level features for establishing correspondence. 
% story
The proposed \textit{deep spectral correspondence} (DSC) determination scheme harnesses the representational power of local feature descriptors to derive a complex high-level global shape representation for matching disparate images. The proposed scheme reasons about correspondence between disparate images using high-level global shape cues derived from low-level local feature descriptors. 
% as a result
Consequently, the proposed scheme enjoys the best of both worlds, i.e., a high degree of invariance to affine parameters such as scale, orientation, viewpoint, illumination afforded by the global shape cues and robustness to occlusion provided by the low-level feature descriptors. While the shape-based component within the proposed scheme infers what to look for, an additional saliency-based component dictates where to look at thereby tackling the noisy correspondences arising from the presence of textureless regions and complex backgrounds. In the proposed scheme, a joint image graph is constructed using distances computed between interest points in the appearance (i.e., image) space. Eigenspectral decomposition of the joint image graph allows for reasoning about shape similarity to be performed {\it jointly}, in the appearance space and eigenspace. 
% what else
Furthermore, a new benchmark dataset consisting of disparate image pairs with extremely challenging variations in scale, orientation, viewpoint, illumination and affine projection parameters and characterized by the presence of complete or partial occlusion of objects, is introduced. The proposed dataset is supplemented with ground truth interest point annotations and is the largest and most comprehensive amongst publicly available image datasets pertaining to the problem of disparate image matching. 
% did it work
The proposed scheme yields state-of-the-art performance in the case of both, coarse-grained shape-based correspondence determination as well as fine-grained point-wise correspondence determination on two existing challenging datasets as well as the newly introduced dataset.

%% Introduction -------------------------------------------------------

\section{Introduction}\label{introduction}

% story for the paper
In the midst of data-driven, annotation-intensive deep learning methods, we propose a learning-free, model-based technique, that reasons about correspondences in the underlying shape space of the scene objects to enable disparate image matching. While applications of mainstream deep learning-based models have experienced tremendous success in several computer vision problems, this success has come at a hefty cost; the cost of gathering and annotating training data is excessive at the very least, and can be quite exorbitant in some cases. This motivates us to revisit learning-free, model-based techniques, especially in the context of correspondence determination for disparate image matching. The proposed \textit{deep spectral correspondence} (DSC) estimation scheme is based on discovery of underlying shape representations using low-level edge-based features and integration of the representational power of high-level shape cues with the robustness of low-level edge features, without recourse to explicit learning. 

%% pipeline fig -------------------------------------------------------

\begin{figure*}
 \centering 
  \graphicspath{{./figures/}}
  \centerline{\includegraphics[width=\textwidth, width=\textwidth+0.5in]{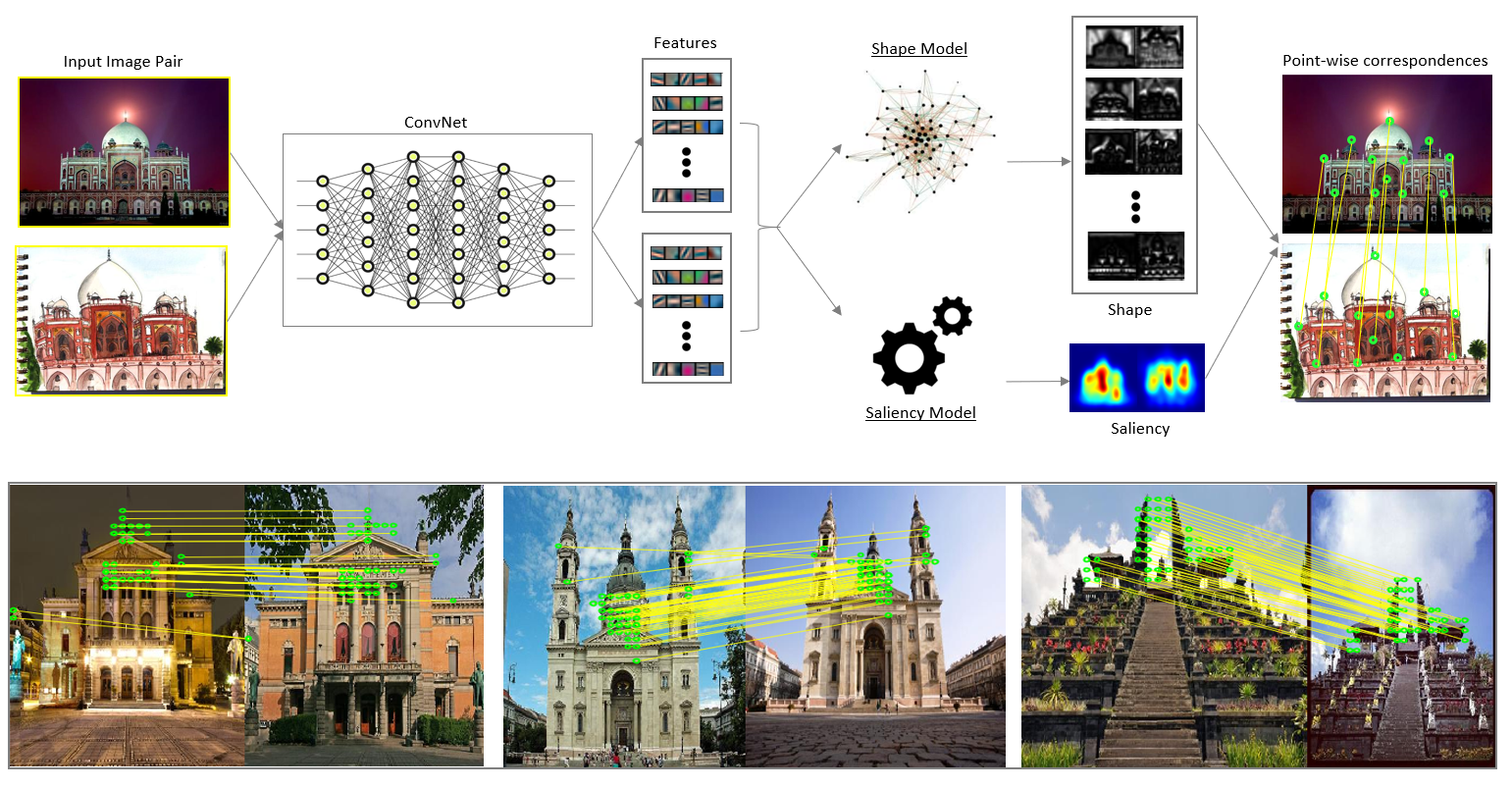}}
  \caption{\footnotesize Pipeline of the proposed deep spectral correspondence (DSC) determination scheme for disparate image matching (top tow) with sample qualitative results (bottom row).}  
  \label{fig:network_architecture}
\end{figure*}

% problem description & its application, challenges, 
Image matching is one of the most fundamental problems in the field of computer vision and, despite significant advances in the field, is still far from being solved. It is imperative to address this problem effectively, since doing so would impact a wide range of computer vision applications such as structure-from-motion (SfM), simultaneous localization and mapping (SLAM), structure-from-stereo (SfS), object localization, fine-grained object categorization, shape-based image retrieval and image registration, just to name just a few.
In spite of the large volume of research over the past couple of decades devoted to tackling this problem, determining a reliable correspondence even between image pairs that exhibit modest variations in viewing conditions, is still immensely challenging~\citep{Bansal13}.

% how we approach and importance
In this paper we tackle a specific subproblem within the larger category of image matching problems, i.e., matching of disparate image pairs. Matching images that are significantly disparate in nature, i.e., images that exhibit extreme variations in viewing conditions characterized by scale, orientation, viewpoint, illumination and affine projection parameters, further complicates the already challenging image matching problem. The formulation of an accurate, efficient and robust solution to the disparate image matching problem is of utmost importance on account to its direct applicability to several real-world problems such as enabling night vision in autonomous cars, generating 3D reconstructions of historical landmarks from internet-scale images to enable photo-tourism, browsing and searching large-scale image repositories, to cite a few.

% why challenging
Appearance similarity measures computed between disparate images using local point-based features exhibit significant inconsistencies on account of the fact that correspondence determination by local feature matching is highly inaccurate~\citep{Bansal13, Mukhopadhyay16}.
To address the above issue, we propose a correspondence determination scheme for matching disparate images that is based on the extraction of a \textit{latent shape representation} from the underlying images. A major drawback of most global or holistic shape representations is their inability to deal with high levels of deformation, articulation and scene occlusion~\citep{Belongie02, Zhu08}. In contrast, local keypoint-based image matching techniques are more robust to high levels of deformation and articulation and instances of partial occlusion~\citep{barnes2009patchmatch,brown2005multi}. However, the inability to reason about shape as a holistic entity, causes most local shape representation techniques to perform poorly when faced with significant variations in the viewpoint and affine projection parameters.

Typically, matching image pairs involves identifying salient regions or keypoints (also referred to as interest points) in the images~\citep{Lowe04} followed by direct correspondence determination between the images using a suitable interest point descriptor-based similarity measure~\citep{trzcinski2013}. In some cases, it is possible to exploit location-based dependencies between the interest points to improve the robustness of the correspondence using model-fitting techniques such as the \textit{random sample consensus} (RANSAC) algorithm~\citep{fischler1981}. 

While the above approach is computationally efficient and often finds practical application in several structure-from-motion (SfM), structure-from-stereo (SfS) and image registration algorithms, it arguably fails to reason about the scene objects or structures at a global level since the entire image is treated as a set of independent interest points~\citep{Bansal13,Mukhopadhyay16}. 

In more recent interest-point-based approaches, reasoning about the scene objects or structures is facilitated by the construction of constellations of known interest points~\citep{fergus2003object} or regions~\citep{felzenszwalb2008discriminatively} to represent scene objects or structures positioned within the image against a common background.

Such reasoning allows one to learn more complex higher-level shape representations~\citep{Belongie02}, which, in turn, allow one to tackle more effectively, instances of partial occlusion, object articulation or object deformation in the underlying images. In recent years, modeling of object shapes has been explored extensively and employed successfully in tasks such as object detection and image segmentation~\citep{Belongie02,Bai07,Zhu08}.

Classical methods for shape matching can be categorized typically based on the granularity of the extracted features. While global feature-based shape matching methods~\citep{Belongie02,Bai07,Zhu08} lack the ability to handle strong articulations and occlusions, their local feature-based counterparts~\citep{barnes2009patchmatch, brown2005multi} fall short of generating optimal results when faced with variations in illumination, viewpoint, scale and orientation. Real world images differ from images of objects captured in a highly controlled environment in that the latter tend to focus on the most prominent scene object. Images captured in highly controlled environments tend to place the most important object under consideration close to the image center, a characteristic which may not necessarily be shared by real-world images.

Most existing shape-based image matching techniques can be classified into two broad categories; (a) techniques that exploit a {\it global} shape-based similarity measure for the underlying objects and, (b)  techniques that compute a measure of correspondence based on matching of {\it local} interest points which is then used to reason about the underlying object shapes. Techniques in the former category follow a top-down approach that implicitly models the holistic shape(s) of the underlying object(s)

whereas techniques in the the latter category use a bottom-up strategy that connects the detected interest points to explicitly reason about the global object shape.

Modeling an object shape explicitly using a set of interest points allows for a straightforward representation and simplistic reasoning about the shape~\cite{felzenszwalb2008discriminatively}. The shape can be conceived as a set of 2D interest points connected by a contour or as a constellation of interest points where the relative positions of the interest points describes the underlying shape~\citep{fergus2003object}. Consequently, shape deformations that can be attributed to variations in viewing conditions and/or articulation, are captured explicitly by reasoning about the interest point positions and their variations~\citep{fergus2003object,felzenszwalb2008discriminatively}. While such a shape representation is reasonable, it entails the learning of a shape prior. Moreover,  representing a complex shape merely as a set of connected points results in an oversimplified shape representation. 

Global representation techniques that model the shape implicitly, typically transform the underlying shape into a lower-dimensional representation. Shape reasoning is then performed primarily in this lower-dimensional representational space. 
The proposed DSC determination scheme does not rely solely on either global shape features or local point-based features; instead, the descriptive power of local interest points is harnessed to generate an implicit global representation of the underlying shape. This implicit global representation is then exploited for the purpose of matching scene structures in disparate image pairs.

 \begin{figure*}
 \centering 
  \graphicspath{{./figures/}}
  \centerline{\includegraphics[width=\textwidth, width=\textwidth+0.5in]{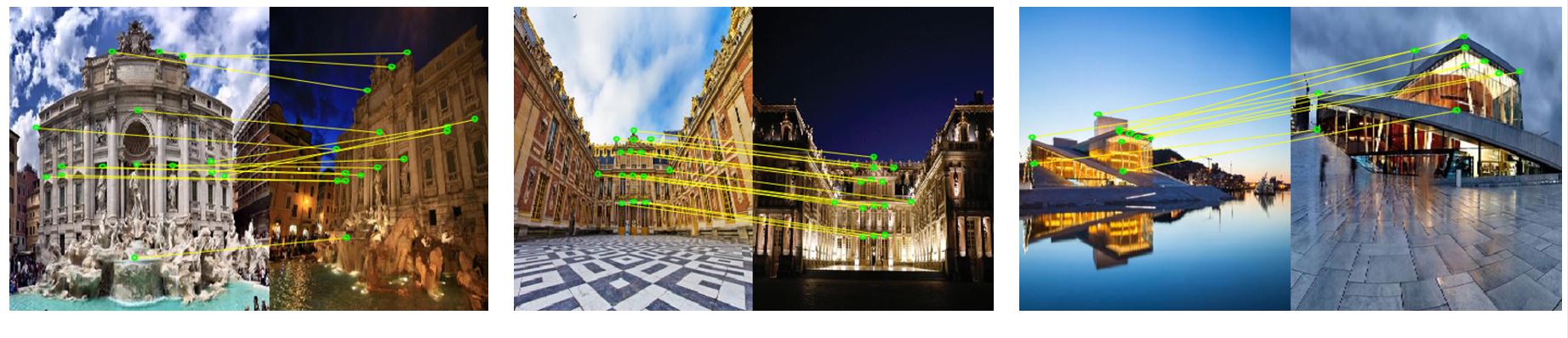}}
  \caption{\footnotesize Example image pairs from the proposed dataset with ground truth annotation of key point correspondences.}
  \label{fig:dataset}
\end{figure*}

In recent times, deep learning approaches have been shown to be very successful in a wide variety computer vision applications. Several recent research works have attempted to leverage the advantages of deep learning for the purpose of matching disparate image pairs. These attempts include the formulation of deep neural networks (DNNs) trained end-to-end as well as extraction of deep-learned features for image matching~\citep{Altwaijry16,Zagoruyko15}. In this paper, we harness the power of deep-learned features, and use them in conjunction with a traditional 2D shape representation, to match disparate images in a suitably defined shape eigenspace. We leverage the sophisticated representation offered by deep-learned features to reason about scene objects in a low-dimensional shape eigenspace obtained by computing the eigenspectrum defined over a suitably defined graph embedding distance space. This low-dimensional representation of image structures has been shown to capture persistent shape cues more effectively than most keypoint-based shape matching approaches~\citep{Bansal13}. 

In our previous work we proposed a partial shape matching technique using \textit{joint spectral embedding} (JSE), where we construct a joint image graph that captures the degree of patch-wise similarity between image structures~\citep{Mukhopadhyay16}. In this paper, we extend our previous work to formulate a saliency-aware and deep learning-based spectral correspondence (DSC) determination scheme, for matching images in shape eigen-space. To this end, we first obtain a low-dimensional representation of the joint image graph via a process of eigenspectral decomposition. This low-dimensional representation is then used to determine shape-based correspondences between image structures in a disparate image pair. Correspondence determination between two image structures in this low-dimensional joint shape eigenspace has two major advantages: (a) the resulting shape cues are more robust to variations in viewing conditions thus yielding a more robust image matching procedure, and (b) the joint representation enables extraction of shape cues that exist across both images, alleviating the burden of matching independently extracted eigenvectors from two distinct eigen representations. Figure~\ref{fig:network_architecture} depicts the overall computational pipeline for the proposed scheme for disparate image matching.

The primary contributions of the paper can be summarized as follows: 
\begin{enumerate}
\item Formulation and implementation of a novel shape- and saliency-aware \textit{deep spectral correspondence} (DSC) determination framework. The proposed framework is based on the principle of joint geometric graph embedding~\citep{Mukhopadhyay16} that  addresses the hitherto unsolved problem of matching disparate image pairs via a process of 2D shape discovery.

\item Introduction of a novel benchmark dataset comprising of highly disparate image pairs. 
It is worthy to note that the proposed benchmark, consisting of over 1000 disparate image pairs, is, to the best of our knowledge, the largest of its kind. Figure~\ref{fig:dataset} shows some sample image pairs from our dataset with annotations of ground truth key point correspondences.

\end{enumerate}

The rest of the paper is organized as follows: We review related work in Section~\ref{sec:related_work} followed by the problem statement in Section~\ref{sec:problem_statement}. In Section~\ref{sec:model}, we present the formulation of the problem, including a formal exposition of the \textit{joint spectral embedding} (JSE) problem (Section~\ref{subsec:JGGE}), incorporation of the saliency term (Section~\ref{subsec:saliency}) and the regularization term (Section~\ref{subsec:regularization}) and formulation of the objective function and optimization procedure for the proposed \textit{deep spectral correspondence} (DSC) determination (Section~\ref{subsec:energy-minimization}). In Section~\ref{sec:dataset}, we introduce and describe a new benchmark dataset \textit{DispScenes} that is designed specifically for the testing and evaluation of disparate image matching algorithms. In Section~\ref{sec:experimental_evaluation}, we detail the performance evaluation metrics for the proposed DSC determination scheme (Section~\ref{subsec:eval_matching}) followed by an experimental comparison of the proposed DSC determination framework to state-of-the-art image matching techniques (Section~\ref{subsec:discussion}). Finally, we conclude the paper in Section~\ref{sec:conclusion_future_work} with an outline of directions for future work.

\section{Related Work}\label{sec:related_work}

\noindent{{\bf Global Shape Representation:}} Conventional global shape matching techniques, which include contour-based and region-based methods~\citep{Belongie02, Shotton08, Kim00, Sebastian01}, compare two shapes by first computing a suitably defined similarity or distance measure between them. The similarity or distance measure is used to formulate a global matching cost function which is then optimized to determine the best match. Contour-based methods exploit primarily the boundary information for matching the underlying shapes. Global contour-based shape matching techniques are based on computation of the shape context measure~\citep{Belongie02}, chamfer distance measure~\citep{Opelt06, Shotton08} and set matching-based contour similarity measure~\citep{Zhu08} followed by optimization techniques based on dynamic programming~\citep{Ravishankar08} and {\it shape skeleton}-based contour matching~\citep{Bai07}. Other relevant works in contour-based global shape matching include the triangle area representation~\citep{Alajlan07} and segment-based shape matching methods such as the shape tree method~\citep{Falzenszwalb07} and the hierarchical procrustes matching procedure~\citep{Mcneill06}.  Global contour-based shape matching methods though capable of capturing the global shape of the object, are unable to handle strong articulation-based deformations of the object shapes. 

Region-based approaches, in contrast, derive global shape descriptors using pixel-level information within an image region bounded by the object shape. Prominent region-based global shape matching methods are based on the computation of invariant moments such as the Zernike moments~\citep{Kim00}. Skeleton-based shape descriptors~\citep{Sebastian01,Sebastian04} have proven better than conventional contour-based methods at capturing shape articulation but their performance deteriorates when dealing with complex shapes due to the absence of region-based descriptors. The general inability of global shape matching techniques to handle shape deformation arising from strong articulation or occlusion of objects (or their parts) has motivated the design of local shape matching techniques.

\vspace{0.2cm}
\noindent{{\bf Local Shape Representation:}} Local shape matching techniques attempt to address the problem of shape deformation in their underlying formulation~\citep{Chen08,Ma11}. Although robust to modest degrees of shape deformation and shape articulation, and capable of providing an accurate measure of local similarity, local shape matching techniques are typically unable to provide a strong global description for accurate shape alignment. Also, most local matching techniques call for prior knowledge of the underlying shape when dealing with the matching of highly articulated object shapes, thus severely limiting their scalability to real-world matching problems~\citep{Chen08, Ma11}.

Another key aspect of shape recovery from local feature points lies in identifying the feature points that are highly reliable and contribute meaningfully to the matching procedure. Feature point correspondences could be improved by identifying and localizing special interest points that ensure the computational stability of the solution for the position, orientation and scale of the matching feature points~\citep{Shi94}. Patch matching techniques attempt to improve feature point correspondences by incorporating multiple views of region patches~\citep{Brown05} and/or mid-level cues from region patches and their nearest-neighboring patches using random sampling~\citep{Barnes09}. However, extreme variations in viewing conditions often encountered in real-world images render the patch matching techniques infeasible.  Since the proposed DSC determination scheme incorporates optimization of the spectral embedding of local geometric features of the joint image graph coupled with region-based regularization and saliency computation, it inherently restricts the extracted features to a suitably determined shape or region within the image resulting in improved correspondences between feature points. 
\vspace{0.2cm}

\noindent{{\bf Spectral Methods for Shape Correspondence: }} Spectral methods employed on the Laplacian of a suitably defined image graph have been proposed in recent research literature, especially in the context of feature clustering and image segmentation~\citep{Arbelaez11}. The proposed technique is motivated by recent work on the use of the eigenspectra of scale-invariant feature transform (SIFT) features~\citep{Lowe04} of a joint image graph as descriptors of image structure~\citep{Bansal13}. It has been shown that the eigenspectral analysis of the joint image graph constructed using dense pixel-level SIFT features extracted from a pair of images, can yield matching features that are robust and persistent across illumination changes~\citep{Bansal13}. In particular, features that encode the extrema of the eigenfunctions of the joint image graph are shown to be stable, persistent and robust across wide range of illumination variations~\citep{Bansal13}. 

\vspace{0.2cm}
\noindent{{\bf Deep Learning for Image Matching:}} In their recent work, \cite{detone2017toward}, \cite{dosovitskiy2015flownet} and \cite{Zagoruyko15} have attempted to harness the representational power of deep learning methods to aid image matching. Deep learning methods for correspondence determination can be categorized as either local interest point-based~\citep{detone2017toward,Zagoruyko15, yi2016lift} or optical flow-based~\citep{dosovitskiy2015flownet, mayer2016large}.
\cite{Zagoruyko15} have explored a variety of convolutional neural network (CNN or ConvNet) architectures to learn a similarity function for unsupervised matching of region patches in images. Furthermore, \cite{detone2017toward} and \cite{simo2015discriminative} have proposed deep learning-based neural architectures for discovery of SIFT-like feature points or region patch representations~\citep{simo2015discriminative} thereby facilitating the incorporation of keypoint prediction architectures into traditional SfM/SLAM pipelines. More recently,~\cite{detone2017superpoint, detone2017toward} have proposed a self-supervised learning architecture for discovery of keypoints from single images. The keypoint discovery process is enabled by warping images using known transformations to generate image pairs that could be used as training data for a supervised learning procedure. \cite{yi2016lift} use  3D scene reconstructions generated using standard SfM techniques as training data for a supervised Siamese-network that learns to predict keypoints in the input images.

Some recent papers, have revisited traditional optical flow computation methods in the context of image matching~\citep{dosovitskiy2015flownet, mayer2016large}. Instead of discovering keypoints in the underlying image, optical flow-based methods regress the correspondences as a continuous function in a data-driven manner. \cite{dosovitskiy2015flownet} have presented an early attempt to formulate optical flow computation as a supervised learning problem which was subsequently improved upon by~\cite{mayer2016large} to predict high-resolution optical flow maps. In more recent work,~\cite{zbontar2016stereo} have explored the use of stereo pairs of image patches~\citep{seki2016patch, mayer2016large} and attempted to leverage the representational 3D features of image patches to improve optical flow computation. Unlike conventional interest point-based methods~\citep{simo2015discriminative, detone2017toward}, optical flow-based methods require very little supervision but need vast amounts of training data in the form of video sequences~\citep{mayer2016large, dosovitskiy2015flownet}. A major drawback of optical flow-based methods is that, they lack the ability to model strong object articulations. Additionally, the correspondences generated by optical flow-based methods are often too noisy to establish reliable point-wise matches.

The proposed DSC estimation technique harnesses the discriminative power of the deep learning features (local) to uncover the underlying latent global shape representation from images, thereby combining both the global and the local shape representations. The use of spectral (eigen) decomposition to represent shapes as opposed to the use of contours or constellations of interest points, naturally allows us to capture the complexity of the shape representations better. While the deep learning models used for feature extraction and saliency computation are pre-trained on other larger datasets, we simply use them for extracting features, and employ no further training procedure for the task of matching images. 
Although the formalism of~\cite{Bansal13} bears some resemblance to the formalism underlying the proposed DSC determination scheme, it is important to note that in the proposed scheme, the partial shape matching is based on features that are not only robust to illumination variations but also to variations in perceived shape geometry resulting from occlusions of object subparts and changes in viewpoint and orientations of objects and/or their local subparts. The proposed technique is also loosely motivated by the partial shape matching technique described by~\cite{Bronstein08}; however, an important difference between the work of~\cite{Bronstein08} and the proposed technique is that the former deals primarily with objects whose surfaces are described by well-defined triangular meshes generated in an artificial environment, whereas the proposed technique tackles the more general and challenging domain of matching objects in real-world images that exhibit significant variability in several imaging and viewing parameters, including cases where the underlying 2D object geometry is not consistent across the images under consideration. 	  

\section{Problem Formulation}\label{sec:problem_statement}

Matching disparate image pairs is a hitherto unsolved computer vision problem, where, for most part, the underlying challenges arise from the extreme variations in viewing and imaging conditions that typify most real-world images. These extreme variations typically stem from variations in scale, orientation, viewpoint, illumination and affine projection parameters accompanied by the presence of complete or partial occlusion of scene objects. The proposed DSC determination scheme explicitly focuses on the problem of matching extremely disparate image pairs where source of the disparateness can be attributed to the aforementioned factors. 

Humans reason about objects or scenes at a more semantic level. Instead of matching points or image patches to establish correspondence, humans are observed to leverage their prior knowledge and ability to extract meaningful high-level (shape) representations, to reason and establish precise correspondences. Inspired by the above observation and the work of~\cite{Bansal13}, in this paper, we propose to extract an \textit{implicit} high-level global shape representation from the input images using standard low-level image features. The resulting shape representation is then used to reason about correspondences between the images. 

Given a disparate image pair, we first extract local image features from the individual input images. These local features are then used to generate a joint graph that embeds the inter-feature distances between the input images in feature space. An implicit high-level global shape representation is then obtained via an eigenspectral decomposition of the joint graph. The global shape representation is shown to span a low-dimensional subspace of the high-dimensional eigenspace of the joint graph. We use the computed global shape representation to establish more precise point-wise correspondences between the images that comprise the disparate image pair. In the context of disparate image matching, a global shape representation has some notable advantages over its low-level feature-based counterpart, since the global shape cues are, in general, more robust to affine transformations and variations stemming from occlusion, deformation and articulation of shapes and changes in scene illumination and viewpoint.

\section{Deep Spectral Correspondence (DSC)}\label{sec:model}

In the proposed approach, we assume that the pair of disparate images being matched contains a dominant object. Also, the images are assumed to exhibit high degrees of variation in scale, orientation, viewpoint,  illumination and affine projection parameters accompanied by the presence of complete or partial occlusion of scene objects and scene structures. The degree of dissimilarity between the image pair ($X$, $Y$) is expressed using a non-negative feature distance function $d: \Sigma^X \times \Sigma^Y \rightarrow$ \begin{math} \mathbb{R}+ \end{math}, where $\Sigma^X$ denotes the appearances cues associated with the regions in image $X$. In our case, each input image is subject to a regular rectangular tessellation, resulting in a lattice of equal-sized rectangular image patches where the center of each patch represents a interest point.  The $i^{th}$ interest point ${K_i}^X  = ({\Sigma_i}^X, {\beta_i}^X )$,  $i=1,2, \ldots, n$ in image $X$ comprises of a location  ${\beta_i}^X$ and its corresponding (local) appearance ${\Sigma_i}^X$ where $n$ is the total number of interest points in image $X$. Let $\mathbf{\varphi}:X \rightarrow Y$  denote the transformation function that maps the interest points in image $X$ to the corresponding interest points in image $Y$. For every interest point $x \in X$,  $\mathbf{\varphi}(x) \in Y$ represents a corresponding unique interest point in image $Y$ with similar local appearance and local geometry. In the continuous case, the corresponding optimization can be formulated as:

\begin{equation}
(X^*, Y^*, \mathbf{\varphi^*}) = \underset{X,Y ; \mathbf{\varphi}:X \rightarrow Y }{\arg\min} \left [ \int_{X \times Y} d(x, \varphi(x)) dx \right ] \label{eq:contOPTIM} 
\end{equation}
The goal of the optimization in eq.~(\ref{eq:contOPTIM}) is to determine the optimal transformation $\mathbf{\varphi^*}$ and the optimal subsets of keypoints (i.e., dominant interest points or  salient regions)  $X^*$ $\subset$  $X$ and $Y^*$ $\subset$  $Y$ that contribute to the transformation.

As mentioned previously, in a discrete setting,  images $X$ and $Y$ are represented as a collection of rectangular image patches centered around the interest points. Hence the optimization of eq.~(\ref{eq:contOPTIM}) can be reformulated as:
\begin{equation}
(X^*, Y^*, \mathbf{\varphi^*})  = \underset{X, Y ; \mathbf{\varphi}: X \rightarrow Y }{\arg\min} \left [ \sum_{X \times Y} d(x, \mathbf{\varphi}(x)) \right ] \label{eq:discOPTIM} 
\end{equation}
In eq.~(\ref{eq:discOPTIM}) we minimize the overall dissimilarity $d(\cdot, \cdot)$ between the corresponding feature points (i.e., interest points) in images $X $ and $Y$  instead of between {\it all} points within the corresponding image patches (i.e., salient regions) in images $X $ and $Y$ . This approximation is justified by an important result in machine learning which states that correct partial point correspondences between two manifolds can be used to infer the complete alignment between all points on the two manifolds~\citep{Ham04}.

In eq.~(\ref{eq:discOPTIM}), the geometric feature distance $d(\cdot, \cdot)$ between two feature points is computed using the  framework described in Section~\ref{subsec:JGGE}. Additionally, we  introduce two regularization terms in the objective/energy function in eq.~(\ref{eq:discOPTIM}). The first regularization term $\mathbf{r}(\cdot,\cdot)$ incorporates a  feature-based (dis)similarity measure between the image patches whereas the second regularization term $\mathbf{s}(\cdot, \cdot)$ embodies a measure of feature saliency of the image patches. The feature (dis)similarity term $\mathbf{r}(\cdot,\cdot)$ encapsulates the global structure of the image regions whereas the saliency-based term $\mathbf{s}(\cdot, \cdot)$  serves to de-emphasize the non-discriminative interest points or regions within the images $X$ and $Y$. Consequently, the final optimization framework is formulated as: 
\begin{equation}
\begin{split}
(X^*, Y^*, \mathbf{\varphi^*})   =  \underset{X, Y ; \mathbf{\varphi}: X \rightarrow Y }{\arg\min} \Bigg [\; \lambda_1 \cdot  \left \{ \sum_{X \times Y} d(x, \mathbf{\varphi}(x))\right \} \\ + \lambda_2 \cdot \mathbf{r}(X, Y) +  \lambda_3 \cdot \mathbf{s}(X,Y) \; \Bigg ] 
\end{split} 
\label{eq:OPTIM} 
\end{equation}
where $\sum_{i=1}^{3} \lambda_i = 1$. 

\subsection{Joint Spectral Embedding (JSE)}
\label{subsec:JGGE}

Spectral analysis of the contents of an image is typically performed on a weighted image graph $G(V,E,W)$ where $V$ is the vertex set, $E$ the edge set and $W$ the affinity matrix associated with the edge set $E$. A vertex $v \in V$ represents a pixel-level appearance descriptor associated with a specific interest point or salient region in the image whereas the edge set $E$ denotes the pair-wise relationships between each pair of vertices in the set $V$, making $G$ a complete graph. The pair-wise relationship denoted by the edge $(v_i, v_j) \in E$  is quantified by a weight $w_{ij} \geq 0$ that encodes the measure of similarity between the pixel-level appearance descriptors associated with vertices $v_i$ and $v_j$. The edge weights are represented using an $n \times n$ affinity matrix $W=[w_{ij}]_{i,j=1,2,...,n}$, where $n = |V|$ is the number of interest points or salient regions in the image. 

The graph formulation above is extended to a joint graph for an image pair $(I_1, I_2)$ as follows: Let $G_1(V_1,E_1,W_1)$ and $G_2(V_2,E_2,W_2)$ be the image graphs for images $I_1$ and $I_2$, respectively. The joint image graph $G(V,E,W)$ is defined by the vertex set $V = V_1 \cup V_2$ and edge set $E =E_1 \cup E_2 \cup (V_1 \times V_2)$ where $V_1 \times V_2$ is the set of edges $(v_i^1, v_j^2)$ for each vertex $v_i^1 \in V_1$ : $i =1, 2, \ldots, n_1$, and each vertex $v_j^2 \in V_2$ : $j =1, 2, \ldots, n_2$,  where $n_i = |V_i|$ is the number of vertices (i.e., interest points or salient regions) in image $I_i$. The resulting affinity matrix $W$ is given by:

\begin{equation}
   W=
  \left[ {\begin{array}{cc}
   W_1 & C \\
   C^T & W_2 \\
  \end{array} } \right]_{(n_1+n_2) \times (n_1+n_2)}   \label{eq:EMBD}
\end{equation}
\noindent where the affinity submatrices $W_1$, $W_2$ and $C$ are defined as follows:
\begin{equation}
(W_i)_{x,y}  =  \exp( -  (\|{f_i(x) - f_i(y)}\|)^2 ) \; {\rm for}\; i = 1,2. 
\end{equation}
\begin{equation}
C_{x,y} =  \exp( -  (\|{f_1(x) - f_2(y)} \|)^2) 
\end{equation}
where $f_i(x)$ and $f_i(y)$ are the pixel-level feature vectors at locations $x$ and $y$ respectively in image $I_i$ and $\| \cdot \|$ is the cosine distance between them.  
% do we need the ^2 - arun2

\begin{figure*}[!ht]
 \centering 
  \graphicspath{{./figures/}}
  \centerline{\includegraphics[width=\textwidth, width=\textwidth+0.5in]{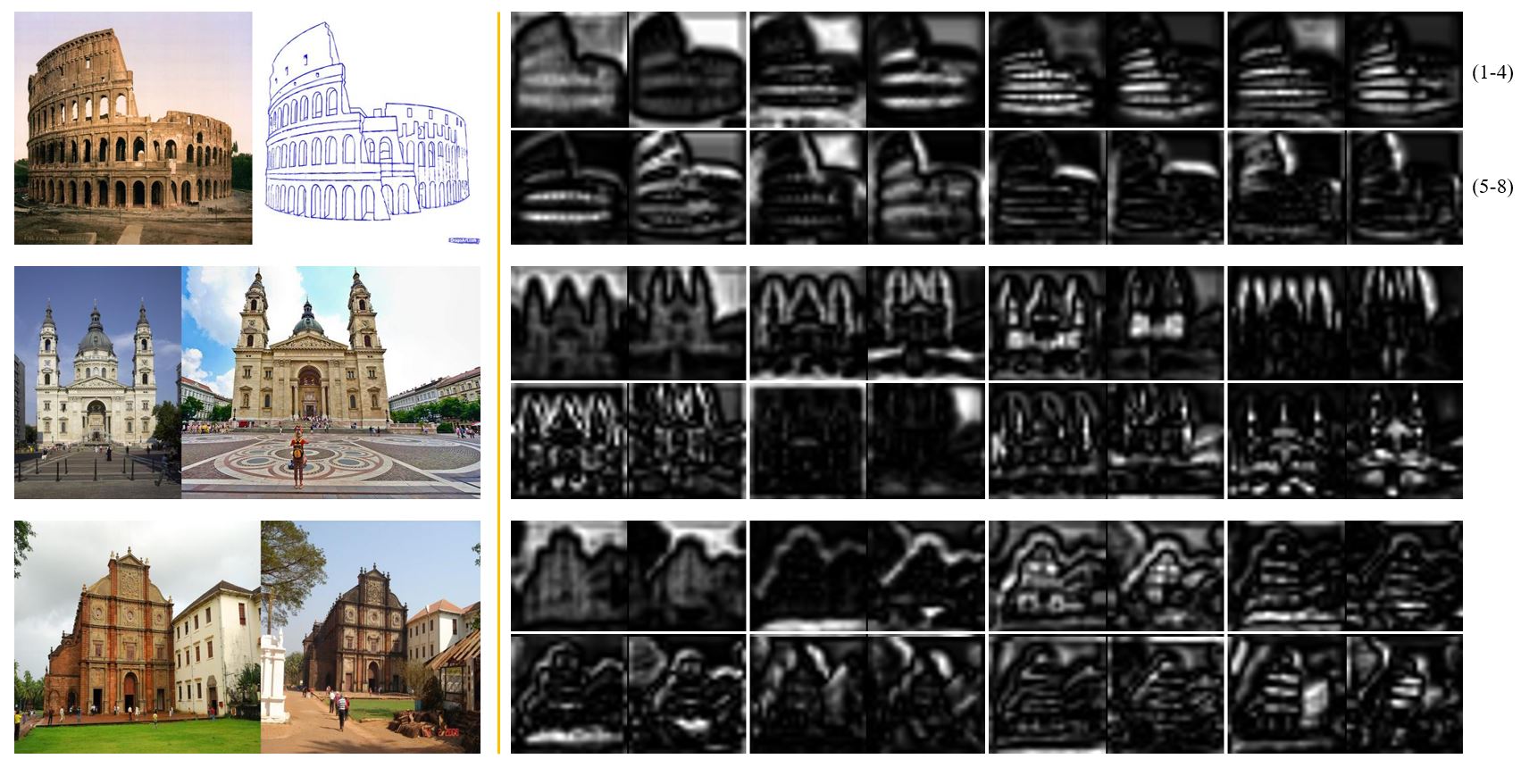}}
  \caption{\footnotesize Visualization of the top-8 eigenvector pairs computed from the joint spectral embedding using the extracted CNN or ConvNet features. Each input image pair is followed by the eigenvector pairs represented as image pairs; the top-(1--4) eigenvector pairs are displayed in the upper half of the row and the top-(5--8) eigenvector pairs are displayed in the lower half of the row corresponding to the input image pair. It can be observed that each eigenvector pair exhibits similar shape properties on account of the joint eigendecomposition.}
  \label{fig:eigenvectors}
\end{figure*}

We compute the {\it joint spectral embedding distance} (JSED) between two pixel-level feature vectors at locations $x$ and $y$ in two different images (comprising the disparate image pair) using the first $m$ non-trivial eigenvectors ($\phi_k$) corresponding to the $m$ smallest non-trivial eigenvalues of the joint graph as follows:
\begin{equation}
d^{2}_{JSED}(x, y) = \sum_{k=1}^m {(\phi_k(x) - \phi_k(y))^2} \label{eq:JSED} 
\end{equation}
\cite{Bansal13} have shown that the joint geometric graph embedding (JSE) procedure ensures very good correspondence between the shapes and distributions of the eigenextrema obtained from the two images being matched. The JSE procedure is shown to reduce the divergence between features derived from the corresponding regions of the two images. This ensures that regions in the two images that exhibit strong correspondence in the JSE space are noted to be in visual agreement with the correspondence results in image space~\citep{Bansal13}. Consequently, the JSED measure $d_{JSED}$ computed in the JSE space is more robust to the variations in imaging and viewing parameters commonly encountered in real-world images than a simple feature distance measure computed in the corresponding feature space. The JSED measure $d_{JSED} (\cdot, \cdot)$ is used as the geometric feature distance $d (\cdot, \cdot)$  in the first term of the energy function in eq.~(\ref{eq:OPTIM}). Figure~\ref{fig:eigenvectors} depicts the top-8 eigenvectors resulting from the joint spectral embedding of some example image pairs from our benchmark dataset using CNN or ConvNet features.

\subsection{Saliency Term}\label{subsec:saliency}

Deep-learned features are global descriptors that are naturally oblivious to saliency. We have observed from our ablation studies that the optimization in eq.~(\ref{eq:discOPTIM}) often gets trapped in a local minimum, specifically on account of interest points that are neither salient nor describable. Unlike handcrafted feature descriptors, such as the scale-invariant feature transform (SIFT)~\citep{Lowe04}, ConvNet (or CNN) features~\citep{simonyan2014very} are not capable of automatically identifying salient interest points; instead, they compute a global description of the underlying image. Consequently, each patch or region in the image is described using a feature descriptor irrespective of its degree of saliency. As a result, image patches lacking in textural features are included in the optimization in eq.~(\ref{eq:discOPTIM}) and weighted to the same degree as salient patches. Inclusion of low-saliency image patches in the optimization in eq.~(\ref{eq:discOPTIM}) results in several mismatches, causing the optimization procedure to get trapped in a local minimum. 

\begin{figure*}
 \centering 
  \graphicspath{{./figures/}}
  \centerline{\includegraphics[width=\textwidth, width=\textwidth+0.5in]{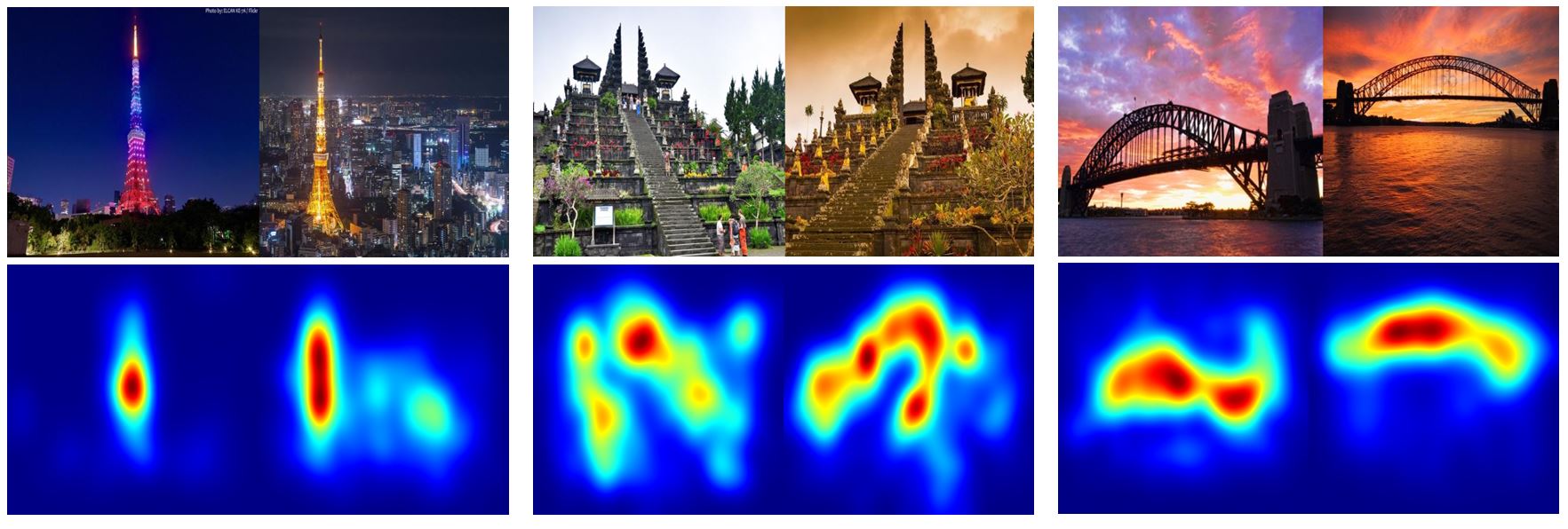}}
  \caption{\footnotesize Saliency computation: input image pairs (top row) and computed saliency map (bottom row).}
  \label{fig:saliency}
\end{figure*}

In order to compute the saliency map for the underlying image when using ConvNet features, we use the scheme proposed by~\cite{Tavakoli17} which assigns a non-negative score (likelihood) to each interest point in order to penalize correspondences between interest points that are not salient. The saliency regularization term is observed to be unnecessary when using features other than deep-learned features. Figure~\ref{fig:saliency} depicts the saliency maps obtained from the sample images in our dataset. The saliency term $\mathbf{s}(X,Y)$ in eq.~(\ref{eq:OPTIM}) keeps track of how salient each region or interest point is in the given image, and  either rewards or penalizes them based on their degree of saliency. Figure~\ref{fig:saliency} depicts the saliency maps for some example image pairs from our benchmark dataset. 

\subsection{Regularization}\label{subsec:regularization}

Owing to the large variations in imaging and viewing conditions coupled with the complexity of the solution landscape, it is possible for the global shape-based reasoning procedure based on the optimization in eq.~(\ref{eq:discOPTIM}) to find itself trapped in a local minimum. To address this shortcoming, we propose to use an auxiliary regularization term, whose primary purpose is to bail the optimization procedure out of a local minimum. We supplement the objective function in eq.~(\ref{eq:discOPTIM}) by the incorporation of a feature (dis)similarity-based regularization term $\mathbf{r}(X,Y)$ that measures the appearance (dis)similarity in feature space between key points in the images $X$ and $Y$.

The feature (dis)similarity-based regularization term $\mathbf{r}(X,Y)$ computes the feature distance between corresponding interest points across the two images. 
The assumption is that, if two points match well in the transformed shape space (i.e., JSE space), they must be proximal in the appearance (feature) space as well. 
The quality of the match between images $X$ and $Y$ in the image pair under consideration is measured using an appropriately defined region-based irregularity function $\mathbf{r}(X',Y')$, where $X' \subset X$ and $Y' \subset Y$ are fixed-sized rectangular image patches centered around the keypoints that are extracted at varying resolutions from the images $X$ and $Y$ respectively. 
We have explored options for measuring the appearance (dis)similarity between the image patches for the purpose of computing the region-based irregularity function; an obvious choice is to use directly the distance in feature space between the appearance descriptors of the corresponding keypoints in the image pairs extracted using a ConvNet. We have also explored using other feature descriptors for measuring the appearance (dis)similarity, such as the difference in the values of the mean pixel intensity (MPI) and the difference of histogram of oriented gradient (HOG) features between the image patches $X'$ and $Y'$.

\begin{algorithm}
\caption{Deep Spectral Correspondence Estimation algorithm}\label{alg:JGGE_Algo}
\small
\begin{algorithmic}[1]

\item Extract pixel-level features from images $I_1$ and $I_2$ belonging to the disparate image pair $(I_1, I_2)$.

\item Compute weighted image graph $G_i(V_i, E_i, W_i)$ for each image $I_i$ using pixel-level features $V_i$ as vertices, pair-wise relationships between vertices $v_k \in V$ and $v_l \in V$ as edges $E_i$ and affinity submatrix $W_i$ which represents the edge weights $(W_i)_{x,y} = \exp( - (\|{f_i(x) - f_i(y)}\|) ^2)$ , where $f_i(x)$ and $f_i(y)$ are, respectively, the pixel-level features at locations $x$ and $y$ in image $I_i$.

\item Compute inter-image weights $C_{x,y} = \exp( - (\|{f_1(x) - f_2(y)} \|)^2)$ between the vertices of $I_1$ and $I_2$. 

\item Compute joint image graph $G(V, E, W)$ where $V = V_1 \cup V_2$, $E =E_1 \cup E_2 \cup (V_1 \times V_2)$ and the resulting affinity matrix is given by $W$ in eq.~(\ref{eq:EMBD}).

\item Compute eigenvectors ($\phi_k$) for the affinity matrix $W$. Use the top $m$ non-trivial eigenvectors corresponding to the $m$ smallest non-trivial eigenvalues of the joint image graph $G$ to compute the joint graph embedding distance using eq.~(\ref{eq:JSED}).

\item Use the proposed multiresolution gradient descent technique
on the multicriteria energy function in eq.~(\ref{eq:OPTIM}) to achieve
region-based matching.

\end{algorithmic}
\normalsize
\end{algorithm}

\subsection{Energy Minimization} \label{subsec:energy-minimization}

In order to compute a partial shape matching using JSE, minimization of the energy function in eq.~(\ref{eq:OPTIM}) is performed using a multiresolution gradient descent technique similar to the one described in~\citep{Bronstein06}. Due to the non-convex nature of the energy function in eq.~(\ref{eq:OPTIM}), conventional gradient descent-based optimization techniques typically converge only to a local optimum. In contrast, multiresolution gradient descent schemes have been observed to exhibit reduced sensitivity to the presence of local optima in the solution space in the context of image registration~\citep{cole-rhodes03, eastman01}. The gradient descent procedure is performed at each level of resolution on equal-sized and overlapping rectangular patches. The optimization in eq.~(\ref{eq:OPTIM}) is performed iteratively, first at a coarsest resolution level and then interpolated to successively finer resolution levels, until convergence is reached at the finest resolution level.

\section{Dataset}\label{sec:dataset}

While the problem of matching disparate images is considered very important in the field of computer vision, surprisingly, there are only a very few dedicated datasets available for testing and evaluation of disparate image matching algorithms. Most of available image matching datasets focus on image patch matching based on determination of patch-wise correspondences~\citep{Chandrasekhar14,Winder07}, while a few deal with image matching across considerable variation in viewpoint, e.g., wide-baseline stereo matching~\citep{Geiger12,Mishkin13,Mishkin15} and variation in illumination~\citep{Hauagge12}. Currently there are no datasets specifically designed for benchmarking of disparate image matching algorithms that account for all types of variations that make disparate image matching a particularly challenging problem. Furthermore, the problem of estimating pair-wise correspondences for fine-grained image matching requires datasets with ground truth annotations of pair-wise point correspondences. Generating such datasets is an extremely laborious process even if the point-wise correspondence annotations across an image pair are sparse.

In this paper, we introduce a new benchmark dataset \textit{DispScenes}, designed specifically with the problem of disparate image matching in mind. The DispScenes benchmark dataset is an extension of our previous benchmark dataset described in~\citep{Mukhopadhyay16}. The DispScenes dataset includes additional image pairs with ground truth annotations of pair-wise point (pixel) correspondences between the images in the image pair. The DispScenes dataset contains 1036 disparate image pairs in total, representing a significant advancement over the previous version that contained only 40 image pairs~\citep{Mukhopadhyay16}.

The disparate image pairs in the DispScenes dataset exhibit dramatic variations in object appearances which arise primarily from affine transformations (scale, translation, rotation and projection), changes in viewpoint, presence of occlusion, and inconsistent ambient illumination. Most image pairs in the DispScenes dataset consist of a single dominant object that is either completely or partially visible. The images in the DispScenes dataset are comprised mostly of architectural structures, monuments and other outdoor scenes. As mentioned previously, the image pairs in the proposed dataset exhibit high levels of variation in illumination (day {\it vs.} night), viewpoint, presence of occlusion as well as other parameters such as age of structures (historic {\it vs.} new) and inclusion of sketches of structures alongside their original counterparts. 

\begin{figure*}[!ht]
 \centering 
  \graphicspath{{./figures/}}
  \centerline{\includegraphics[width=\textwidth, width=\textwidth+0.5in]{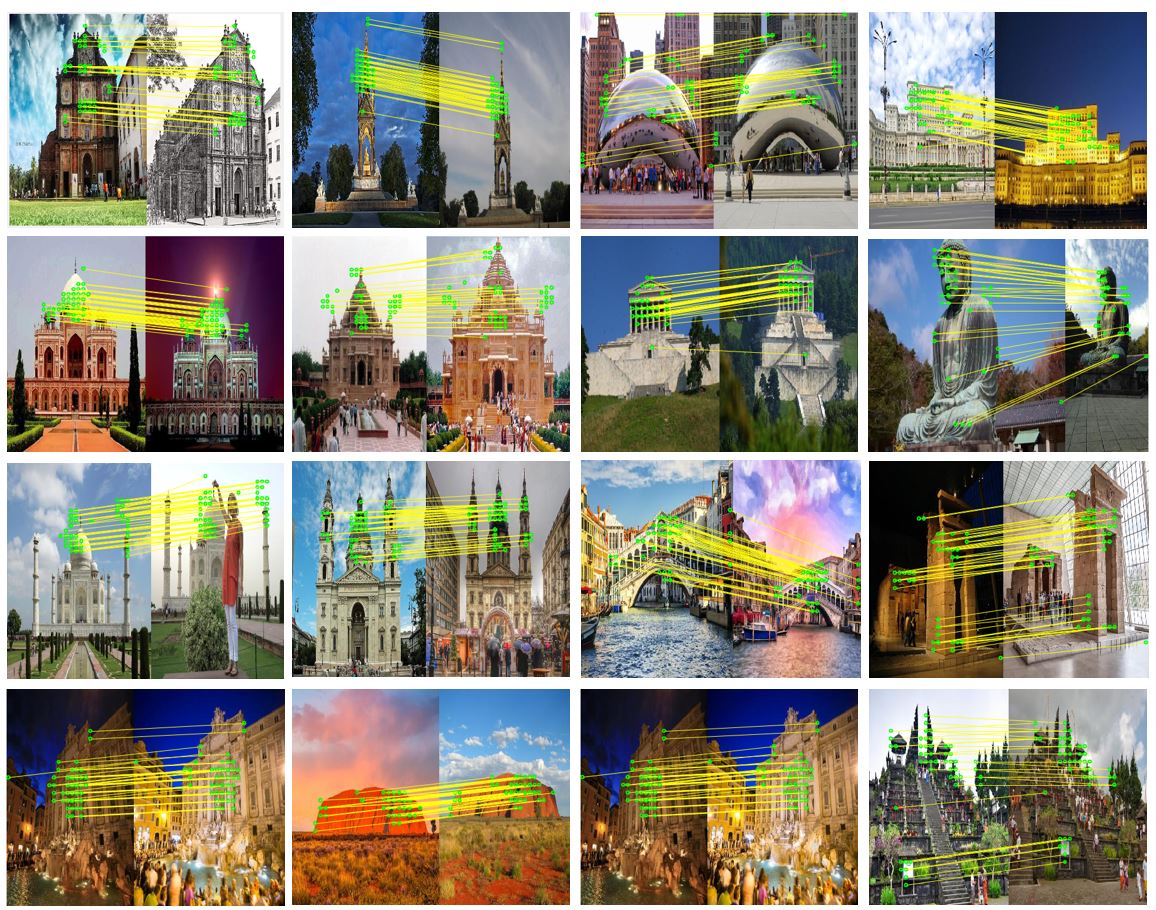}}
  \caption{\footnotesize Visualization of the \textit{good} qualitative results of the deep spectral correspondence (DSC) determination scheme on image pairs from the proposed benchmark dataset. For each image pair, the source image is on the left and the target image on the right.} \label{fig:qual_good}
\end{figure*}

The DispScenes benchmark dataset is, by far, the largest amongst the image matching datasets specifically designed for benchmarking of disparate image matching algorithms. Moreover, unlike previous datasets~\citep{Hauagge12, Mukhopadhyay16}, the DispScenes dataset also provides manually annotated, ground truth pair-wise keypoint correspondences between the images in an image pair. The number of pair-wise keypoint annotations between the images in an image pair ranges between 5 and 14. The dataset allows for the testing and evaluation of image matching algorithms based on estimation of fine-grained, pair-wise point correspondences. We also provide the homogeneous matrix computed using the pair-wise point correspondences which can then be used to generate additional keypoints via a process of interpolation. We identify the image pairs in which the homogeneous matrix (estimated using the ground truth keypoints) fails to find accurate correspondences, on account of the high extent of variation in viewpoint, and categorize them as {\it difficult}, whereas the rest of image pairs are categorized as {\it easy}. Furthermore, the above categorization is manually ascertained by visual verification of the matching error between annotated ground truth keypoints and the points interpolated using the homogeneous matrix. Using the above procedure, out of the 1036 image pairs in the DispScenes benchmark dataset, we categorize 362 images as {\it easy} and 674 images as {\it difficult}.

\begin{figure*}[!ht]
 \centering 
  \graphicspath{{./figures/}}
  \centerline{\includegraphics[width=\textwidth, width=\textwidth+0.5in]{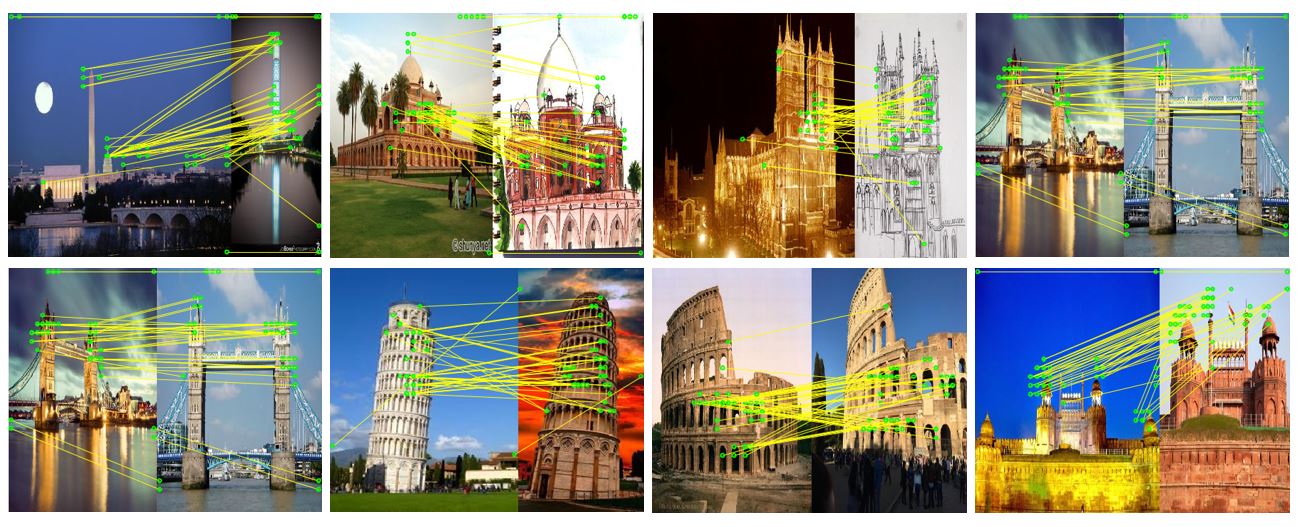}}
  \caption{\footnotesize Visualization of the \textit{bad} qualitative results of the deep spectral correspondence (DSC) determination scheme on image pairs from the proposed benchmark dataset. For each image pair, the source image is on the left and the target image on the right.} \label{fig:qual_bad}
\end{figure*}

\begin{figure*}[!ht]
 \centering 
  \graphicspath{{./figures/}}
  \centerline{\includegraphics[width=\textwidth, width=\textwidth+0.5in]{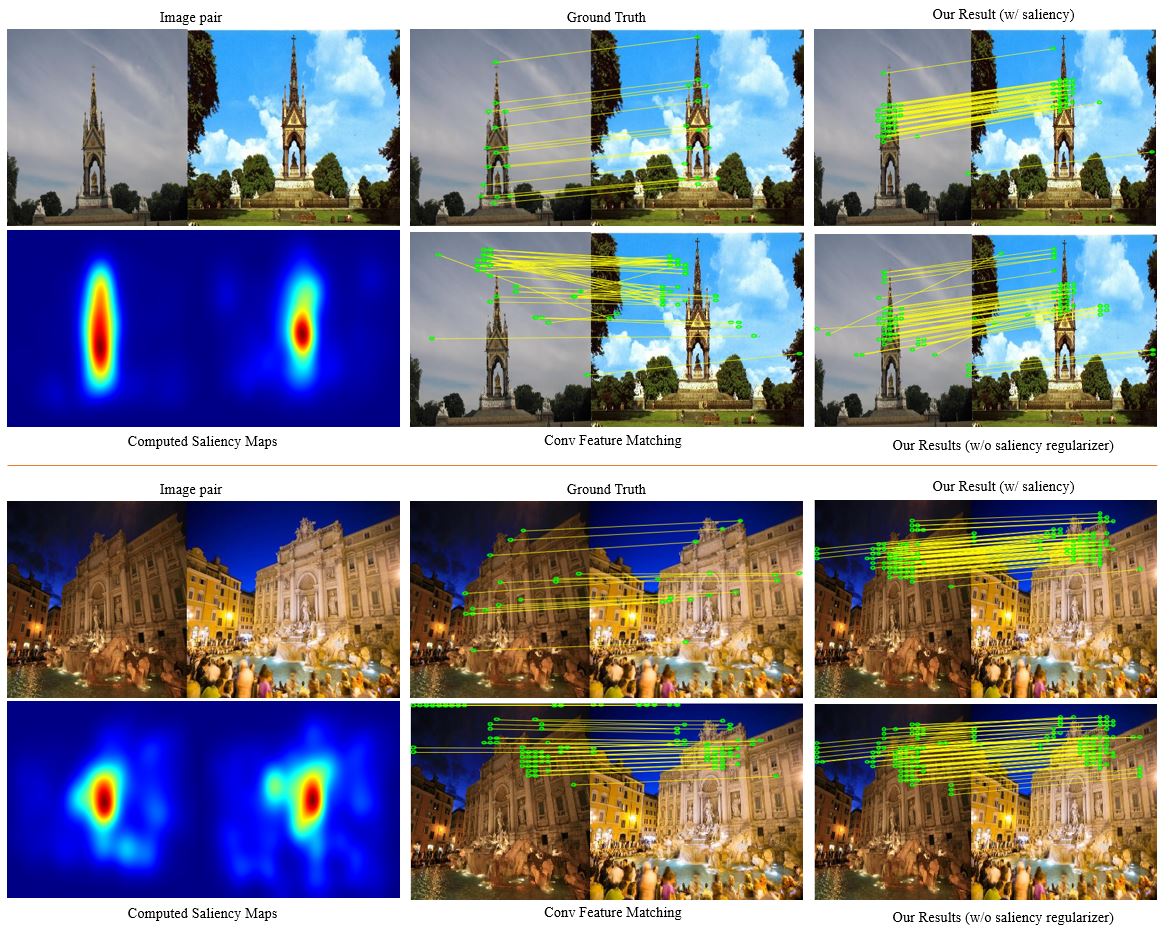}}
  \caption{\footnotesize Comparison of qualitative results. Top row: input image pair, ground truth correspondences and results of the proposed DSC determination scheme with the saliency-based regularization term. Bottom row: saliency map, correspondences estimated using CNN or ConvNet features and results of the proposed DSC determination scheme without the saliency-based regularization term.}  
  \label{fig:qual_1}
\end{figure*}

\begin{table*}[t]
\caption{\footnotesize Comparison of standard interest point-based image matching techniques:  SIFT~\citep{Lowe04}, SURF~\citep{Bay06}, FAST~\citep{Rosten06}, ORB~\citep{Rublee11}, GB~\citep{Berg01} to the MSER-FEM technique~\citep{Bansal13}, variants of the proposed joint spectral embedding (JSE) using GB features, and the proposed deep spectral correspondence (DSC) determination technique, on the datasets of~\citep{Mukhopadhyay16} and~\citep{Hauagge12}. The \textit{mean relevance score} (MRS) and \textit{false positive rate} (FPR) are used as performance measures. High values of MRS and low values of FPR are desirable. The term MPI denotes \textit{mean pixel intensity} whereas the term NR denotes \textit{no regularization})} 
\vspace{-0.1 in}
 \label{tab:Compare1} 
\centering
\fontsize{7pt}{7pt}\selectfont
    \begin{tabular}{ c | c | c | c | c | c | c | c | c | c | c}
    \toprule
   \multicolumn{8}{l}{\textbf{Results ($R_1$) on dataset of}~\citep{Mukhopadhyay16}} \\ \hline
    \ & SIFT & ORB & SURF & FAST & GB & MSER-FEM & JSE-GB-MPI & JSE-GB-HOG & JSE-GB-NR & DSC\\ \midrule    
    MRS & 0.49 & 0.63 & 0.59 & 0.40 & 0.36 & 0.68 & \textbf{0.82} & 0.79 & 0.60 & \textbf{0.91}\\ \hline
    FPR & 0.48 &  0.34 & 0.37 & 0.54 & 0.61 & 0.31 & \textbf{0.17} & 0.19 & 0.40 & \textbf{0.07}\\ \toprule
    \multicolumn{7}{l}{\textbf{Results ($R_1$) on dataset of}~\citep{Hauagge12}} \\ \hline
   \ & SIFT & ORB & SURF & FAST & GB & MSER-FEM & JSE-GB-MPI & JSE-GB-HOG & JSE-GB-NR & DSC\\ \midrule    
    MRS & 0.66 & 0.68 & 0.76 & 0.50 & 0.73 & 0.72 & 0.90 & \textbf{0.91} & 0.87 & \textbf{0.96}\\ \hline
    FPR & 0.31 & 0.28 &  0.21 & 0.47 & 0.24 & 0.27 & 0.10 & \textbf{0.08} & 0.11 & \textbf{0.03}\\  \bottomrule
    \end{tabular}
 \normalsize
\end{table*}

\begin{table*}[t]
\caption{\footnotesize Comparison of variants of the proposed \textit{joint spectral embedding} (JSE) technique, using different feature descriptors such as SIFT~\citep{Lowe04}, ORB~\citep{Rublee11}, GB~\citep{Berg01} in conjunction with MPI-based and HOG-based regularization described in Section~\ref{subsec:regularization} to the proposed DSC determination scheme. The \textit{mean relevance score} (MRS) and \textit{false positive rate} (FPR) are used as performance measures. High values of MRS and low values of FPR are desirable.}
 \label{tab:Compare2} 
\centering
\fontsize{7pt}{7pt}\selectfont
    \begin{tabular}{ c | c | c | c | c | c | c | c}
    \toprule
   \multicolumn{5}{l}{\textbf{Results on the dataset of}~\citep{Mukhopadhyay16}}  \textbf{based on the performance metric $R_1$}\\ \hline
    \ & JSE-SIFT-MPI & JSE-SIFT-HOG  & JSE-ORB-MPI & JSE-ORB-HOG &  JSE-GB-MPI & JSE-GB-HOG & DSC\\ \midrule
    MRS & 0.74 & 0.73 & 0.78 & 0.78 & \textbf{0.82} & \textbf{0.79} & \textbf{0.91}\\ \hline
    FPR & 0.26 &  0.27 & 0.21 & 0.22 & \textbf{0.17} & \textbf{0.19} & \textbf{0.07}\\ \toprule
    \multicolumn{5}{l}{\textbf{Results on the dataset of}~\citep{Hauagge12} \textbf{based on the performance metric $R_1$}} & \\ \hline
   \ &  JSE-SIFT-MPI & JSE-SIFT-HOG  & JSE-ORB-MPI & JSE-ORB-HOG &  JSE-GB-MPI & JSE-GB-HOG & DSC\\ \midrule    
    MRS & 0.75 & 0.74 & 0.82 & 0.83 & \textbf{0.90} & \textbf{0.91} & \textbf{0.96}\\ \hline
    FPR & 0.24 &  0.25 & 0.17 & 0.16 & \textbf{0.10} & \textbf{0.08} & \textbf{0.03}\\ \bottomrule
    \end{tabular}
 \normalsize
\end{table*}

\begin{figure*}[!ht]
 \centering 
  \graphicspath{{./figures/}}
  \centerline{\includegraphics[width=\textwidth, width=\textwidth+0.5in]{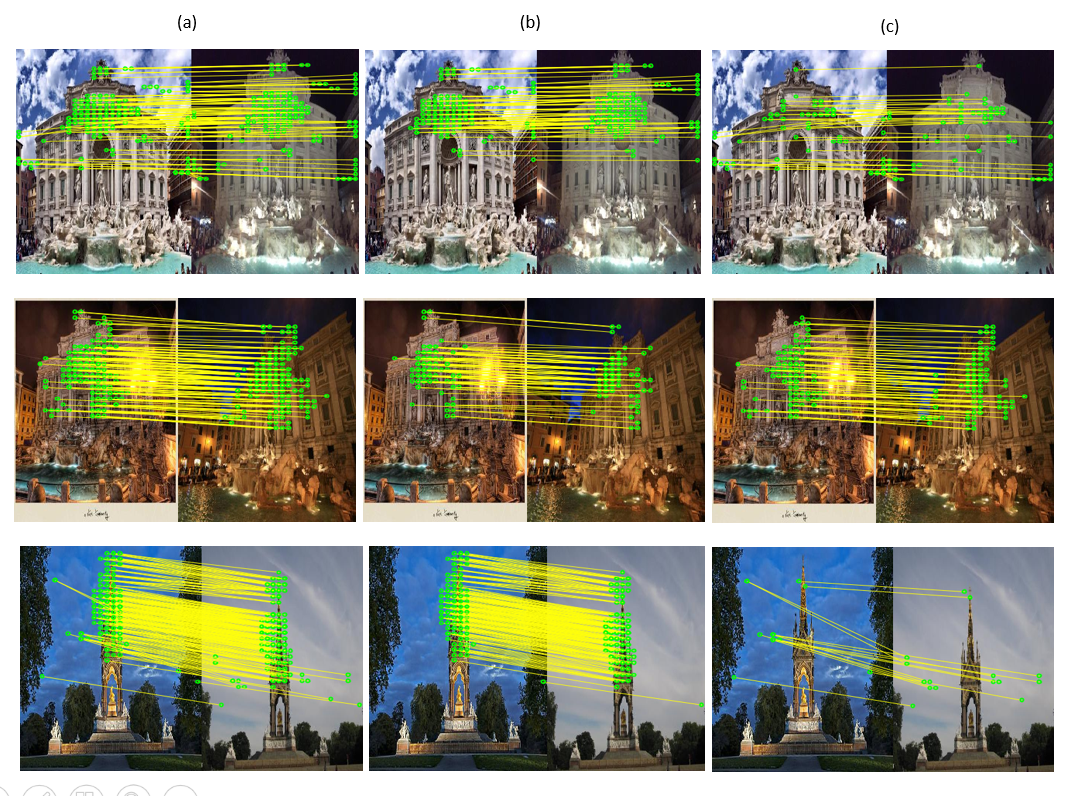}}
  \caption{\footnotesize Visualization of qualitative results: (a) correspondences resulting from the proposed DSC determination scheme (b) {\it valid} correspondences that lie within a $15 \times 15$ window with respect to the ground truth, (c) {\it invalid} correspondences that lie outside $15 \times 15$ window with respect to the ground truth.}  
  \label{fig:qual_2}
\end{figure*}

\begin{table*}[tbp]
\footnotesize
\caption{\footnotesize Comparison of variants of the proposed DSC scheme to the image matching scheme that uses CNN or ConvNet features in isolation (i.e., without joint spectral embedding) in the context of point-wise correspondence estimation on the proposed dataset. Rows 1--4 report the values of the $R_2$ metric averaged over the image pairs in the dataset for four values of scale or granularity ($10 \times 10$, $\ldots$, $40 \times 40$). Each entry in rows 1--4 reports the $R_2$ metric values in the form of $X/Y$ where $X$ is the $R_2$ metric averaged over the image pairs in the dataset deemed to be \textit{easy} whereas $Y$ is the $R_2$ metric averaged over the image pairs in the dataset deemed to be \textit{difficult}. The mean average error (MAE) is reported on row 5 (lower values for MAE implies better matching performance). Saliency and feature-based regularization terms are denoted as $\mathbf{s}$ and $\mathbf{r}$ respectively.}
 \label{tab:Compare3} 
\centering
\scriptsize
    \begin{tabular}{ c | c | c | c | c | c }
    \toprule
   \multicolumn{6}{l}{\textbf{Results on the proposed benchmark dataset {\it DispScenes} based on the performance metric $R_2$}} \\ \hline
    \ & {DSC complete} & DSC {\it(w/o $\mathbf{r}$)} & DSC {\it(w/o $\mathbf{s}$)} & DSC {\it(w/o $\mathbf{s}$ {\rm and} $\mathbf{r}$)} & {ConvNet features}  \\ \hline    
    {10 $\times$ 10} & { \bf 41.83 / 21.09 } & 39.44 / 20.27  & 37.63 / 19.92 & 35.45 / 19.61 & 17.51 / 13.29 \\ 
    {20 $\times$ 20} & { \bf 59.27 / 32.96 } & 56.76 / 31.07 & 54.71 / 30.43 & 50.03 / 29.88 & 22.59 / 19.18 \\ 
    {30 $\times$ 30} & { \bf 65.37 / 36.19 } & 62.44 / 34.53 & 59.02 / 33.10 & 55.78 / 32.57 & 27.96 / 20.03 \\
    {40 $\times$ 40} & {\bf  68.86 / 37.85 } & 65.03 / 35.91  & 64.81 / 34.23 & 59.39 / 33.45 & 32.74 / 20.42 \\ 
    {MAE}            & {\bf  47.93 } & 54.88  & 57.49  & 67.14 & 110.28 \\ 
    \bottomrule
    \end{tabular} 
\end{table*}

\begin{table*}[tbp]
\footnotesize
\caption{\footnotesize Comparison of the complete DSC determination scheme (DSC complete) to the image matching schemes that uses the CNN/ConvNet features and the SIFT features~\citep{Lowe04} in isolation, the MSER-FEM scheme~\citep{Bansal13}, the JSE-GB-MPI scheme~\citep{Mukhopadhyay16} and the Superpoints scheme~\citep{detone2017superpoint} in the context of point-wise correspondence estimation on the proposed dataset. Rows 1--4 report the values of the $R_2$ metric averaged over the image pairs in the dataset for four values of scale or granularity ($10 \times 10$, $\ldots$, $40 \times 40$). Each entry in rows 1--4 reports the $R_2$ metric values in the form of $X/Y$ where $X$ is the $R_2$ metric averaged over the image pairs in the dataset deemed to be \textit{easy} whereas $Y$ is the $R_2$ metric averaged over the image pairs in the dataset deemed to be \textit{difficult}. The mean average error (MAE) is reported on row 5 (lower values for MAE implies better matching performance).}
 \label{tab:Compare4} 
\centering
\scriptsize
    \begin{tabular}{ c | c | c | c | c | c | c}
    \toprule
    
   \multicolumn{7}{l}{\textbf{Results on the proposed benchmark dataset {\it DispScenes} based on the performance metric $R_2$}} \\ \hline
    \ & {DSC complete} & {ConvNet features} & {SIFT features} & {MSER-FEM} & {JSE-GB-MPI} & {Superpoints}\\ \hline    
    {10 $\times$ 10} & {\bf 41.83 / 21.09 } & 17.51 / 13.29 & 18.52 / 8.50 &  21.32 / 10.28 & 23.33 / 13.62 & 27.48 / 23.01\\ 
    {20 $\times$ 20} & {\bf 59.27 / 32.96 } & 22.59 / 19.18 & 23.25 / 11.06 & 24.87 / 12.83  & 26.68 / 15.07 & 38.14 / 26.31\\ 
    {30 $\times$ 30} & {\bf 65.37 / 36.19 } & 27.96 / 20.03 & 26.02 / 11.97 & 27.18 / 13.03  & 29.54 / 17.92 & 45.55 / 27.49\\
    {40 $\times$ 40} & {\bf 68.86 / 37.85 } & 32.74 / 20.42 & 27.60 / 12.64 & 29.02 / 14.77 & 32.11 / 18.26 & 49.63 / 27.89\\ 
    {MAE}            & {\bf 47.93 } &  110.28 & 151.52  & 134.59  & 116.17 & 86.61 \\ 
    \bottomrule
    \end{tabular} 
\end{table*}

\section{Experimental Evaluation}\label{sec:experimental_evaluation}

We demonstrate the performance of the proposed DSC-based image matching scheme on existing datasets~\citep{Mukhopadhyay16} and~\citep{Hauagge12} and on the proposed DispScenes benchmark dataset. The datasets in~\citep{Mukhopadhyay16} and~\citep{Hauagge12} are similar to the proposed dataset but considerably smaller in size, comprising of 40 and 46 disparate image pairs, respectively. We report the matching accuracy of the proposed JSE image matching scheme for varying granularity of correspondences; from coarse-grained region-based correspondence to fine-grained point-wise correspondence. 

For the proposed dataset, we use bounding boxes of varying scales $s_j$ where $j \in \{10,20,30,40\}$, denoting bounding boxes of sizes $10 \times 10$, $20 \times 20$,  $30 \times 30$ and $40 \times 40$. A correspondence is deemed valid at a particular scale $s_j$, if the corresponding point determined by the image matching scheme lies within a bounding box (of the appropriate size) centered at the ground truth corresponding point. In contrast to the proposed DispScenes benchmark dataset, the datasets in~\citep{Mukhopadhyay16} and~\citep{Hauagge12} do not contain annotations for point-wise ground truth correspondences. Consequently, when evaluating the proposed DSC-based image matching scheme on the datasets in~\citep{Mukhopadhyay16} and~\citep{Hauagge12}, we are constrained to use region-based (coarse) correspondences using the ground truth annotations provided by~\cite{Mukhopadhyay16}.

\subsection{Evaluation Metrics}\label{subsec:eval_matching}

To evaluate the performance of the proposed DSC-based image matching algorithm on the datasets in~\citep{Mukhopadhyay16} and~\citep{Hauagge12}, we use the coarse region-based correspondences determined by the image matching algorithm. The region-based correspondences are used to compute a {\it relevance score} $R_1$, which is an {\it intersection-over-union} (IoU) measure computed for each matched image pair as follows: 
\begin{equation}
R_1=\frac{TPR}{TPR+TNR+FPR} \label{eq:RELEV}
\end{equation}
\noindent where, 
\begin{eqnarray}
TPR & = & (X' \cap X'_g) \cup (Y' \cap Y'_g) \\
TNR & = & (X'_g - (X' \cap X'_g)) \cup (Y'_g - (Y' \cap Y'_g)) \\
FPR & = & (X' - (X' \cap X'_g)) \cup (Y' - (Y' \cap Y'_g))
\end{eqnarray}
\noindent
The parameters $X'_g$ and $Y'_g$ denote the ground truth bounding boxes enclosing the dominant object in images $X$ and $Y$ respectively. The parameters $X'$ and $Y'$ denote the corresponding regions in images $X$ and $Y$ determined by the proposed DSC-based image matching algorithm. The parameters {\it TPR}, {\it TNR} and {\it FPR} denote the true positive rate, true negative rate and false positive rate respectively. The relevance score $R_1$ in eq.~(\ref{eq:RELEV}) provides a measure of the coarse region-based correspondence between the images, and is useful when ground truth annotations of fine point-wise correspondences are not available, as is the case for the datasets in~\citep{Mukhopadhyay16} and~\citep{Hauagge12}. We compute the {\it mean relevance score} (MRS) as the average of the $R_1$ scores over all image pairs in the dataset.

Since the proposed DispScenes benchmark dataset provides manually annotated ground truth point-wise correspondences, we use an alternative measure $R_2$ to evaluate the performance of the proposed DSC-based image matching algorithm on the proposed dataset. Given ground truth point-wise correspondences $(p_i, p_j) \in \mathcal{C}$, where $\mathcal{C}$ is the set of ground truth point-wise correspondences and $p_i$ and $p_j$ are corresponding points in the source and target images respectively, the alternative measure $R_2$ is computed as: 
\begin{equation}
\begin{array}{cc}
{R}_{2} = \frac{1}{|\mathcal{C}|}\sum_{(p_i, p_j) \in \mathcal{C}} \varepsilon(p_i, p_j), \quad \varepsilon(p_i, p_j)=\bigg \{\begin{array}{cc} 1 \quad {\rm if} \quad \delta_{i,j} \leq \tau  \\ 0 \quad {\rm otherwise,}\end{array} \\ \\
{\rm where} \quad \delta_{i,j} = \frac{1}{2}\bigg[\;||(p_i - p'_i)||^2 + ||(p_j - p'_j)||^2 \;\bigg]^{\frac{1}{2}}
\end{array}
\label{eq:R2}
\end{equation}
In eq.~(\ref{eq:R2}),  $(p'_i, p'_j) \in \mathcal{C'}$ are the 2D point-wise correspondences estimated by the proposed DSC-based image matching algorithm such that $p'_i$, and $p'_j$ lie within a $\vartheta$-pixel neighborhood of points $p_1$ and $p_2$ in the source and target images respectively. In our experiments we set $\vartheta=5$. The parameter $\delta_{i,j}$ represents the squared correspondence error between the estimated and ground truth 2D point-wise correspondences $(p'_i, p'_j) \in \mathcal{C'}$ and $(p_i, p_j) \in \mathcal{C}$ respectively where $|| \cdot ||$ denotes the Euclidean norm and $\tau$ is a suitably defined threshold. In our experiments, in order to capture the accuracy of the estimated point-wise correspondences at varying levels of granularity, we report results for multiple values for $\tau$. In our experiments we used the values $\tau$ = \{10, 20, 30, 40\}. Based on the formulation in eq.~(\ref{eq:R2}), higher values of $R_2$ imply more accurate point-wise correspondence determination between the source image and target image. In addition, we also compute the \textit{mean average error} (MAE) metric to represent the average of the continuous error between the ground truth positions of the interest points and their estimated positions over all image pairs in the dataset.

\begin{figure*}[!ht]
 \centering 
  \graphicspath{{./figures/}}
  \centerline{\includegraphics[width=\textwidth, width=\textwidth+0.5in]{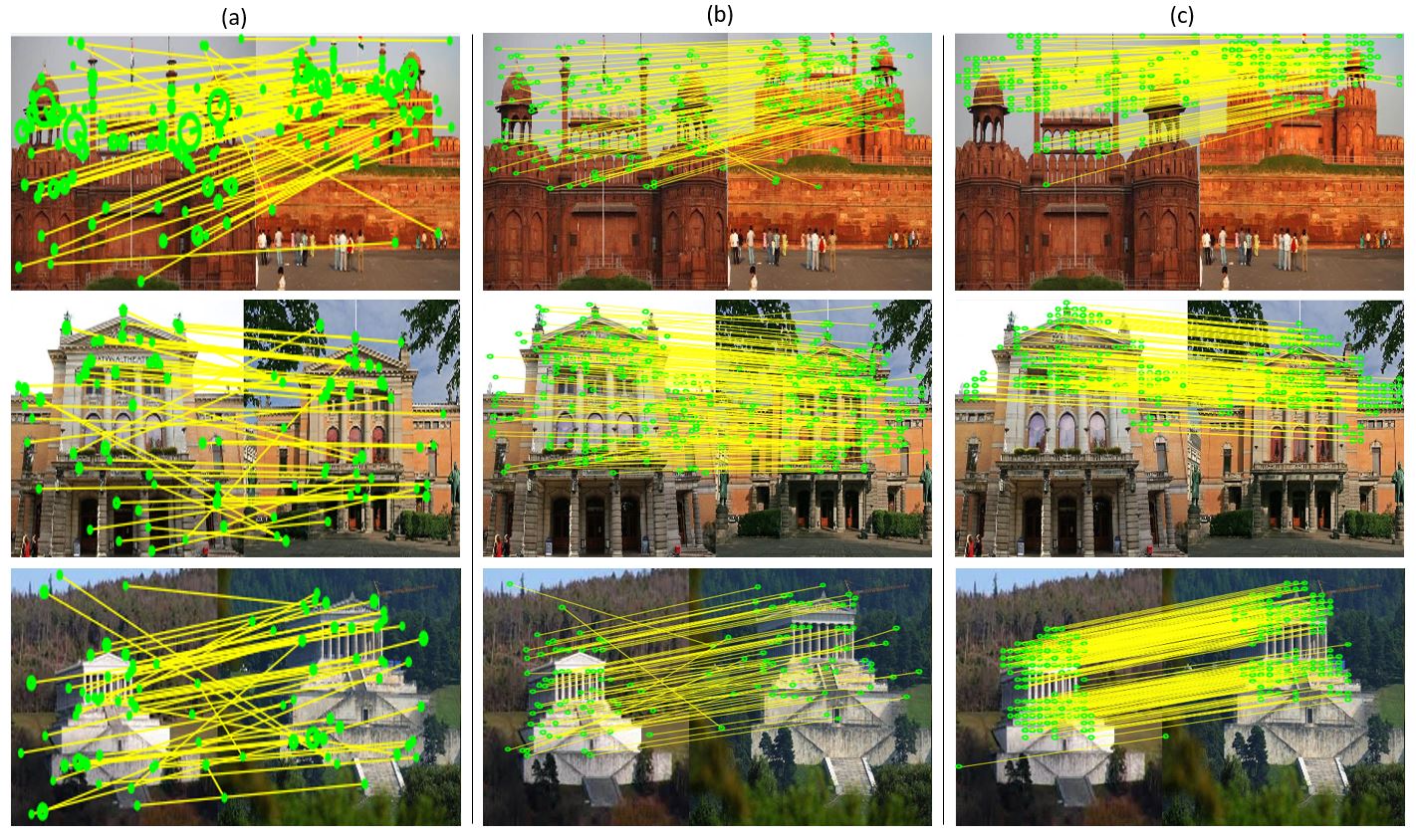}}
  \caption{\footnotesize Comparison of qualitative results: (a) correspondence determination using  SIFT~\citep{Lowe04} feature matching (b) correspondence determination by matching of Superpoints~\citep{detone2017superpoint}, and (c) correspondence estimation using the proposed DSC determination scheme.}
    \label{fig:qual_3}
\end{figure*}

\subsection{Discussion of Results}\label{subsec:discussion}

As described in the previous section, we use two different metrics $R_1$ and $R_2$ to quantify the performance of correspondence determination. The metric $R_1$ is a measure of coarse region-based correspondences (i.e., coarse-grained matching) whereas the metric $R_2$ is indicative of how well the point-wise correspondences are rendered (i.e., fine-grained matching). The metric $R_1$ is used in the case of datasets in~\citep{Mukhopadhyay16} and~\citep{Hauagge12} for which the ground truth point-wise correspondences are not available and the mean $R_1$ score values (i.e., mean relevance score (MRS) values) are tabulated in Tables~\ref{tab:Compare1} and~\ref{tab:Compare2}. The metric $R_2$ is used in the case of the proposed DispScenes benchmark dataset in which the ground truth point-wise correspondences are annotated and the fine-grained point-wise matching scores are reported in Tables~\ref{tab:Compare3} and~\ref{tab:Compare4}. 

While the goal of the proposed shape-aware optimization scheme described in eq.~(\ref{eq:OPTIM}) is to establish correspondence via recovery of the underlying object shapes in highly cluttered images, the sophisticated deep-learned feature descriptions computed using pre-trained CNN or ConvNet models, give the proposed scheme an unfair advantage. While we use deep-learned features as our primary feature descriptor, we also deem it appropriate to demonstrate the power of the proposed joint spectral embedding (JSE) scheme using conventional hand-crafted feature descriptors such as the \textit{scale invariant feature transform} (SIFT)~\citep{Lowe04}, \textit{oriented FAST and oriented BRIEF} (ORB)~\citep{Rublee11} and \textit{geometric blur} (GB)~\citep{Berg01}, along with simple regularizers based on feature distances, such as the \textit{histogram of oriented gradients} (HOG) feature distances~\citep{Dalal05},  or simply the difference between the \textit{mean pixel intensity} (MPI) values of image patches surrounding the keypoints instead of the more sophisticated regularizers described in Section~\ref{subsec:regularization}. 

In Table~\ref{tab:Compare1}, we compare the performance of the proposed DSC determination scheme with the performance of image matching techniques that use standard features such as SIFT~\citep{Lowe04}, \textit{speeded up robust features} (SURF)~\citep{Bay06}, \textit{features from accelerated segment test} (FAST)~\citep{Rosten06}, ORB~\citep{Rublee11} and GB~\citep{Berg01} in isolation (i.e., without joint spectral embedding). We also compare the performance of the proposed DSC determination scheme with the performance of several variants of the proposed JSE scheme such as JSE using GB features and an MPI-based regularizer (JSE-GB-MPI), JSE using GB features and a HOG-based regularizer (JSE-GB-HOG) and JSE using GB features without any regularizer (JSE-GB-NR). We also compare the performance of the proposed DSC determination scheme with the performance of the \textit{maximally stable extremal region-based feature-ellipse matching} (MSER-FEM) scheme described in~\citep{Bansal13}. We report results for the benchmark datasets in~\citep{Mukhopadhyay16} and~\citep{Hauagge12} and use the mean relevance score (MRS) (i.e., the mean value of the $R_1$ metric) and the false positive rate (FPR) as the performance metrics. 

From the results in Table~\ref{tab:Compare1}, it is evident that image matching techniques that use conventional hand-crafted features such as SIFT, SURF, FAST, ORB and GB in isolation (i.e., without joint spectral embedding) do not yield good results (i.e., low MRS values and high FPR values). The proposed JSE scheme with GB features and MPI- and HOG-based regularization (i.e., JSE-GB-MPI and JSE-GB-HOG) performed significantly better than the standard feature-based image matching techniques and the MSER-FEM scheme~\citep{Bansal13}. To evaluate the effectiveness of the regularizer, we also evaluated the performance of the proposed JSE scheme with GB features but without the regularization term (i.e., JGGE-GB-NR) on both datasets. The omission of the regularizer can observed to result in deterioration of the results for datasets. Most importantly, the proposed DSC determination scheme that uses CNN- or ConvNet-derived features is observed to outperform the shape matching techniques that use conventional hand-crafted features in isolation and the and the MSER-FEM scheme~\citep{Bansal13}. 

In Table~\ref{tab:Compare2}, we compare the performance of the proposed DSC determination scheme with variants of the proposed JSE scheme with conventional hand-crafted features such as, SIFT features and MPI- and HOG-based regularization (i.e., JSE-SIFT-MPI and JSE-SIFT-HOG), ORB features and MPI- and HOG-based regularization (i.e., JSE-ORB-MPI and JSE-ORB-HOG) and GB features and MPI- and HOG-based regularization (i.e., JSE-GB-MPI and JSE-GB-HOG) on both benchmark datasets, i.e.,~\citep{Mukhopadhyay16} and \citep{Hauagge12}. Among the variants of the proposed JSE scheme with conventional hand-crafted features, the GB features were observed to outperform all other features. This could be attributed to the inherent ability of GB to focus on feature points of dominant objects that aids in identification of salient region clusters within the image as opposed to just salient keypoints obtained by SIFT~\citep{Lowe04} or SURF~\citep{Bay06}. While the incorporation of a regularization term clearly resulted in improved performance, MPI-based regularization performed better than HOG feature-based regularization in the case of the benchmark dataset in~\citep{Mukhopadhyay16} whereas in the case of the benchmark dataset in~\citep{Hauagge12}, HOG feature-based regularization performed better. This suggests that in simpler image datasets such as~\citep{Hauagge12} where the image disparities could be attributed primarily to variations in viewpoint (to a reasonable extent), texture-based regularization (where HOG features are used to characterize texture) fares well. In contrast, more complex datasets such as~\citep{Mukhopadhyay16}, where the sources of image disparity are more varied, are better served by intensity-based regularization. The proposed DSC determination scheme can be observed to outperform all variants of the JSE scheme with conventional hand-crafted features and  MPI- and HOG-based regularization. 
 
In Tables~\ref{tab:Compare3} and~\ref{tab:Compare4}, we report the performance of the proposed DSC determination scheme in the context of fine-grained point-wise correspondence estimation, for which we leverage the availability of ground truth keypoint annotations in the proposed DispScenes benchmark dataset. In Table~\ref{tab:Compare3}, rows 1--4 report the values of the $R_2$ metric averaged over the image pairs in the dataset for four values of scale or granularity ($10 \times 10$, $\ldots$, $40 \times 40$). Each entry in rows 1--4 of  Table~\ref{tab:Compare3}, reports the $R_2$ metric values in the form of $X/Y$ where $X$ is the $R_2$ metric averaged over the image pairs in the dataset deemed to be \textit{easy} whereas $Y$ is the $R_2$ metric averaged over the image pairs in the dataset deemed to be \textit{difficult}. In addition, we report the mean average error (MAE) on row 5. In Table~\ref{tab:Compare3}, we compare the performance of various variants of the DSC determination scheme such as the complete DSC determination scheme outlined in eq.~(\ref{eq:OPTIM}) comprising of the regularization term ${\bf r} (X, Y)$ and the saliency term ${\bf s} (X, Y)$, the DSC determination scheme without the term ${\bf r} (X, Y)$, the DSC determination scheme without the term ${\bf s} (X, Y)$ and the DSC determination scheme without both terms ${\bf r} (X, Y)$ and ${\bf s} (X, Y)$. In addition we also include the performance of the image matching scheme that uses the CNN or ConvNet features in isolation (i.e., without joint spectral embedding). As is evident from the results in Table~\ref{tab:Compare3}, the complete DSC determination scheme outperforms all the other variants of the DSC determination scheme and the image matching scheme that uses CNN or ConvNet features in isolation in terms of both the $R_2$ metric and the MAE metric. The complete DSC determination scheme exhibits the highest average $R_2$ metric values for both easy and difficult cases across all scales and also yields the lowest overall MAE values. The criterion for categorization of image pairs in the proposed dataset as \textit{easy} or \textit{difficult} is detailed in Section~\ref{sec:dataset}.

In Table~\ref{tab:Compare4}, we compare the performance of the complete DSC determination scheme to the performance of the following: (a) the image matching scheme that uses the CNN or ConvNet features in isolation, (b) the image matching scheme that uses the SIFT features~\citep{Lowe04} in isolation, (c) the MSER-FEM scheme~\citep{Bansal13}, (d) the JSE-GB-MPI scheme~\citep{Mukhopadhyay16} and the \textit{self-supervised interest point detection and description} (Superpoint) scheme~\citep{detone2017superpoint}. The overall format of Table~\ref{tab:Compare4} is identical to that of Table~\ref{tab:Compare3}. From the results in Table~\ref{tab:Compare4} it is evident that complete DSC determination scheme significantly outperforms the other schemes in terms of both, the $R_2$ metric values averaged over both easy and difficult cases across all scales and also yields the lowest overall MAE values. 

Figure~\ref{fig:qual_good} depicts the qualitative results of the DSC determination scheme that are deemed to be \textit{good} (i.e., where the corresponding mean average error (MAE) values $\leq $ a predefined threshold value), whereas Figure~\ref{fig:qual_bad} depicts the qualitative results of the DSC determination scheme that are deemed to be \textit{bad} (where the corresponding MAE values $ > $ the predefined threshold value) on example image pairs from the proposed benchmark dataset. 
Figure~\ref{fig:qual_1} depicts the qualitative results of the proposed DSC determination technique on example image pairs from the proposed benchmark dataset with and without the saliency-based regularization term and compares them with the ground truth correspondences and the correspondences obtained using CNN or ConvNet features in isolation (i.e., without joint spectral embedding). The corresponding saliency maps are also depicted in Figure~\ref{fig:qual_1}. The qualitative results show that the proposed DSC determination technique with the inclusion of the saliency-based regularization term yields results that are closest to the ground truth correspondences. Figure~\ref{fig:qual_2} depicts the valid and invalid correspondences resulting from the proposed DSC determination technique on example image pairs from the proposed benchmark dataset. A computed correspondence is deemed valid if lies within a window of predetermined size ($15 \times 15$ in our case) centered around the closest ground truth corresponding points. Figure~\ref{fig:qual_3} compares the qualitative results of the proposed DSC determination scheme on example image pairs from the proposed dataset with those obtained using SIFT features~\citep{Lowe04} in isolation (i.e., without joint spectral embedding) and Superpoint features~\citep{detone2017superpoint}. The proposed DSC determination scheme can be visually observed to yield fewer false correspondences. 

While the proposed DSC determination scheme (which exploits deep-learned ConvNet features) was observed to exhibit superior results, the CNN or ConvNet features when used in isolation (i.e., without joint spectral embedding) in a conventional feature-based image matching scheme did not perform well (Tables~\ref{tab:Compare3} and~\ref{tab:Compare4}). 
Based on the results in Tables~\ref{tab:Compare1}--~\ref{tab:Compare4} we can conclude that the improved performance of the proposed DSC determination scheme can be primarily attributed to the joint spectral embedding and the proposed shape-aware optimization in eq.~(\ref{eq:OPTIM}). However, improving the quality of the features in the joint spectral embedding and the optimization of the energy function in eq.~(\ref{eq:OPTIM}), e.g., by replacing SIFT or GB features with deep-learned CNN or ConvNet features, does improve the overall performance of the correspondence determination scheme significantly. On the other hand, replacing SIFT or GB features with deep-learned CNN or ConvNet features in a conventional feature-based image matching scheme (that does not employ spectral embedding) does not result in a significant improvement in performance. 

As the ablation study in Table~\ref{tab:Compare3} shows, the proposed DSC determination scheme owes it superior performance to all the following factors: (a) the quality of the features used, (b) the spectral embedding (c) the exploitation of saliency, (d) the use of shape-based regularization and (e) effective optimization of the energy function in eq.~(\ref{eq:OPTIM}). In our experiments, values of the parameters $\{\lambda_1,\lambda_2,\lambda_3\}$, i.e., parameters that determine the relative influence of the shape, regularization and saliency terms in the optimization of the energy function in eq.~(\ref{eq:OPTIM}), are set to $\{$0.75, 0.10, 0.15$\}$ respectively. These values for $\{\lambda_1,\lambda_2,\lambda_3\}$ are learned by maximizing the point-wise correspondence accuracy in the validation set. 

\section{Conclusion and Future Work}\label{sec:conclusion_future_work}

In this paper, we proposed a novel optimization framework for matching disparate images using joint graph embedding. In the proposed framework low-level image cues are used to build high-level shape hypotheses, to reason about image similarity in the shape space. 
The proposed framework incorporates shape saliency and geometry in a deep spectral correspondence (DSC) determination scheme capable of matching highly disparate image pairs. 
By harnessing the representational power of local low-level feature descriptors to derive a complex high-level global shape representation, the proposed DSC determination scheme was shown to be capable of matching image pairs that exhibited high degrees of variation in scale, orientation, viewpoint, illumination and  affine projection parameters, and were also accompanied by the presence of textureless regions and complete or partial occlusion of scene objects. Furthermore, in this paper we also introduced a new benchmark dataset consisting of disparate image pairs with extremely challenging variations in scale, orientation, viewpoint, illumination and affine projection parameters and characterized by the presence of complete or partial occlusion of objects. The proposed dataset, supplemented with ground truth interest point annotations, is the largest and most comprehensive amongst publicly available image datasets pertaining to the problem of disparate image matching. The proposed DSC determination scheme was shown to outperform other state-of-the-art methods, achieving reliable image matching in noisy, real-world images, while circumventing the challenges posed by the degree of variations in viewing conditions. We demonstrated the performance of the proposed method, both qualitatively and quantitatively, on two existing challenging datasets as well as the newly introduced dataset. 

In this paper, the CNN or ConvNet was used only in the feature extraction phase, i.e., to obtain the feature-level descriptors of the keypoints. In future, we plan to extend the proposed DSC embedding scheme to a CNN or ConvNet trained in a data-driven manner in an end-to-end fashion. We also plan to extend the proposed DSC embedding framework to tackle problems dealing with surface-from-motion (SfM), image-based rendering and simultaneous localization and mapping (SLAM). 

\section{Acknowledgments}
\noindent
This research was supported in part by the National Science Foundation under Grant No.
1442672 to Dr. Suchendra M. Bhandarkar. Any opinions, findings, and conclusions or recommendations expressed in this material are those of the author(s) and do not necessarily reflect the views of the National Science Foundation.

\bibliographystyle{ACM-Reference-Format}
\bibliography{refs}

\end{document}